\documentclass[conference]{IEEEtran}
\IEEEoverridecommandlockouts
\usepackage{cite}
\usepackage{graphicx}
\usepackage{overpic}
\usepackage{amsmath,amssymb,amsfonts}
\usepackage{algorithmic}
\usepackage{graphicx}
\usepackage{textcomp}
\usepackage{xcolor}
\usepackage{algorithm}
\usepackage{algorithmic}
\usepackage{amsmath}
\usepackage{xcolor}
\usepackage{booktabs}
\usepackage{graphicx}
\usepackage[table,xcdraw]{xcolor}
\usepackage{hyperref}
\usepackage{tabularray}
\usepackage{graphicx}
\usepackage{subcaption}
\usepackage{caption}
\usepackage{stfloats}
\usepackage{makecell}
\usepackage{booktabs}
\usepackage{multirow}
\usepackage{etoolbox}
\usepackage{orcidlink}
\usepackage[numbers]{natbib}
\usepackage{multirow}
\usepackage{amssymb}
\usepackage[table,xcdraw]{xcolor}
\usepackage{amsthm}
\usepackage{amsmath,amssymb,amsthm}

\def\BibTeX{{\rm B\kern-.05em{\sc i\kern-.025em b}\kern-.08em
    T\kern-.1667em\lower.7ex\hbox{E}\kern-.125emX}}
\begin{document}

\title{KD-MARL: Resource-Aware Knowledge Distillation in Multi-Agent Reinforcement Learning}

\author{
\IEEEauthorblockN{
Monirul Islam Pavel\IEEEauthorrefmark{1}\orcidlink{0000-0000-0000-0000},
Siyi Hu\IEEEauthorrefmark{2}\orcidlink{0000-0000-0000-0000},
Muhammad Anwar Ma'sum\IEEEauthorrefmark{1}\orcidlink{0000-0000-0000-0000},
}

\IEEEauthorblockN{
Mahardhika Pratama\IEEEauthorrefmark{1}\IEEEmembership{Senior Member, IEEE}\orcidlink{0000-0000-0000-0000},
Ryszard Kowalczyk\IEEEauthorrefmark{1}\IEEEauthorrefmark{3}\orcidlink{0000-0000-0000-0000},
Zehong Jimmy Cao\IEEEauthorrefmark{1}\orcidlink{0000-0000-0000-0000}
}

\IEEEauthorblockA{\IEEEauthorrefmark{1}School of Computer Science and Information Technology, Adelaide University, Australia}
\IEEEauthorblockA{\IEEEauthorrefmark{2}School of Electrical Engineering, Computing and Mathematical Sciences, Curtin University, Australia}
\IEEEauthorblockA{\IEEEauthorrefmark{3}Systems Research Institute, Polish Academy of Sciences, Poland}

\IEEEauthorblockA{Corresponding author: monirulislam.pavel@adelaide.edu.au}
}


\maketitle

\begin{abstract}
Real-world deployment of multi-agent reinforcement learning (MARL) systems is fundamentally constrained by limited compute, memory, and inference time. While expert policies achieve high performance, they rely on costly decision cycles and large-scale models that are impractical for edge devices or embedded platforms. Knowledge distillation (KD) offers a promising path toward resource-aware execution, but existing KD methods in MARL focus narrowly on action imitation, often neglecting coordination structure and assuming uniform agent capabilities. We propose resource-aware Knowledge Distillation for Multi-Agent Reinforcement Learning (KD-MARL), a two-stage framework that transfers coordinated behavior from a centralized expert to lightweight, decentralized student agents. The student policies are trained without critic, relying instead on distilled advantage signals and structured policy supervision to preserve coordination under heterogeneous and limited observations. Our approach transfers both action-level behavior and structural coordination patterns from expert policies with supporting heterogeneous student architectures, allowing each agent’s model capacity to match its observation complexity, which is crucial for efficient execution under partial or limited observability along with limited onboard resources. Extensive experiments on SMAC and MPE benchmarks demonstrate that KD-MARL achieves high performance retention while substantially reducing computational cost. Extensive experiments across standard multi-agent benchmarks show that KD-MARL retains over $90\%$ of expert performance while reducing computational cost by up to $28.6\times$ FLOPs. The proposed approach achieves expert-level coordination and can be preserved through structured distillation, enabling practical MARL deployment across resource-constrained onboard platforms.

\end{abstract}

\begin{IEEEkeywords}
Knowledge Distillation, MARL, Resource-Aware Learning,  Edge Intelligence, Compression, Knowledge Transfer
\end{IEEEkeywords}

\section{Introduction}

Multi-agent reinforcement learning (MARL) enables coordination among autonomous agents across domains including robotics \cite{wong2023deep}, distributed sensing \cite{gronauer2022multi}, and autonomous systems. Despite successes, deploying MARL in resource-constrained environments embedded systems, satellite networks remains challenging due to stringent requirements: low latency, limited memory, and real-time responsiveness.

Unlike computer vision or language modeling where compression targets network parameters, MARL's computational burden stems from the \textit{decision-making cycle} rather than model size \cite{yang2024beyond, jiang2019multi}. Continuous policy updates, non-stationary observations, and decentralized coordination substantially increase training and inference costs \cite{nekoei2023dealing}. High-capacity MARL models achieve excellent cooperation but are computationally expensive and memory-intensive \cite{de2021constrained}, limiting deployment in on-board or distributed systems where efficiency is critical.

Knowledge distillation addresses these constraints by transferring behavioral and structural knowledge from high-capacity \textit{teacher} models to lightweight \textit{student} agents \cite{liu2024fine,park2019relational}. Unlike standard RL relying on sparse or delayed environmental rewards, KD leverages supervised expert trajectories, enabling faster convergence and improved stability. In MARL, this distills multiple knowledge forms: \textit{action distributions} serving as soft targets to reduce inference complexity; \textit{coordination dependencies} capturing inter-agent optimization; and \textit{value structure} stabilizing distributed decision-making \cite{yang2025multi}. These representations allow students to emulate expert decisions whilst requiring fewer floating-point operations, enhancing efficiency and reducing latency.

Existing MARL distillation approaches remain limited. Many focus solely on policy imitation without transferring coordination patterns \cite{tseng2022offline, gao2021knowru}, whilst others assume homogeneous agents with identical observations \cite{zhong2024heterogeneous, bao2022recent} unrealistic for practical applications where agents differ in sensing capabilities and operate under partial or noisy observations \cite{gronauer2022multi, wong2023deep}. Moreover, most works neglect the compression-efficiency trade-off, often yielding reduced policy diversity and suboptimal coordination when scaled down \cite{liu2025survey, xu2025survey}.

\begin{figure}[t]
    \centering
    \includegraphics[width=0.5\textwidth]{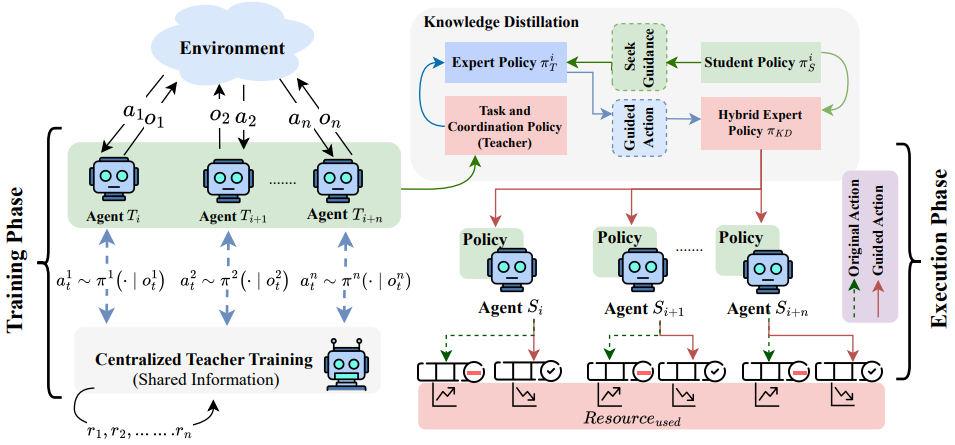}
    \caption{\small Overview of proposed framework for teacher-student based CTDE inspired knowledge distillation in MARL, where the teacher learns expert coordination during training and guides lightweight student policy with near-expert performance \& reduced resource cost.}
    \label{FIG:intro}
    \vspace{-3mm}
\end{figure}

Reformulating cooperative MARL as single-agent RL by treating joint state-action spaces as one virtual decision-maker leads to centralized training and centralized execution (CTCE) \cite{zhao2022ctds,foerster2016learning}. However, this encounters scalability challenges as joint spaces grow exponentially with agent count. MARL also faces partial and heterogeneous observations: agents perceive environments differently based on roles or locations, requiring models accommodating varying input complexity \cite{zhong2024heterogeneous}. Our framework introduces heterogeneous student architectures, assigning smaller models to agents with simpler inputs and larger models to those with richer observations, aligning model expressiveness with input complexity \cite{liu2025survey, hu2023teacher} to avoid unnecessary computational overhead.

We propose KD-MARL, a two-stage knowledge distillation framework enabling multi-agent deployment under strict computational and memory constraints via centralized training, decentralized execution (CTDE) (Fig.~\ref{FIG:intro}). First, a high-capacity expert policy is trained using standard MARL algorithms; second, coordination knowledge is distilled into ultra-lightweight students for efficient operation on limited hardware \cite{tseng2022offline}. A composite distillation objective integrates policy imitation, coordination structure preservation, and RL feedback \cite{xu2025survey}, enabling students to retain expert cooperation whilst adapting to resource limitations and partial observability. The framework supports heterogeneous architectures, aligning each agent's capacity with observation complexity to minimize redundant computation \cite{zhong2024heterogeneous, hu2023teacher}. Our contributions are:
\begin{itemize}
    \item  Introduce KD-MARL, a two stage teacher-student framework that transfers coordinated behavior from a centralized MAPPO expert to lightweight decentralized agents.
    \item Design novel distillation loss combining action-policy fidelity, preservation for students pattern, and coordination under limited heterogeneous observations.

    \item Propose teacher-guided advantage distillation, where GAE-based advantage targets computed from the frozen expert critic replace environment-driven value learning, ensuring stable critic-free policy optimization.

    \item Achieve near-expert performance retention while reducing computational cost by up to $28.6\times$ in MPE and $11.7\times$ in SMAC in terms of FLOPs per episode, with corresponding improvements in inference-time throughput.

\end{itemize}

\section{Related Works}






In deep RL, knowledge distillation enables compressed models to maintain decision-making capabilities of larger counterparts. For multi-agent systems, this addresses inherent MARL scalability challenges. Early work focused on policy transfer under centralized training with decentralized execution. Czarnecki et al. \cite{czarnecki2019distilling} demonstrated distillation across heterogeneous action spaces, whilst Gao et al. \cite{gao2021knowru} proposed KnowRU for structured knowledge reuse. However, these methods require high-quality teacher policies and assume full observability, limiting applicability in partially observable environments.

Recent work addresses these limitations through complementary approaches. Double Distillation Network \cite{ddn2025} incorporates internal and external knowledge signals to improve coordination and exploration in sparse reward settings. For scenarios prohibiting online interaction, offline methods \cite{offline2022} enable distillation from static datasets, though balancing compression with policy expressiveness remains challenging.

Computational efficiency has motivated integrating distillation with network pruning. Liu et al. \cite{liu2023modelcompression} introduced RL-guided compression dynamically removing redundant parameters under teacher supervision. Dan et al. \cite{dan2024pdd} extended this with unified pruning and distillation. Domain-specific applications emerged, such as Chen et al.'s portfolio management system \cite{chen2023portfolio} implementing role-aware knowledge transfer for dynamic task allocation. Ensemble approaches \cite{czarnecki2019distilling} leverage multiple teachers but increase training complexity and may reduce policy diversity.
Despite advances, current methods exhibit significant limitations. CTDS \cite{zhao2022ctds} distills Q-values without model compression or relational modeling between agents. CTPDE \cite{pei2025policy} achieves policy transfer but incurs high computational costs and produces homogeneous behaviors. Transformer-based approaches \cite{tseng2022offline} offer feature-level distillation but require specialized architectures poorly generalizing to standard MARL algorithms. PTDE \cite{chen2024ptde} introduces auxiliary modules for personalized representations, increasing model complexity. MAST \cite{hu2024value} reduces computational load through pruning but lacks mechanisms ensuring behavioral consistency with teacher policies.


\section{Methodology}

\begin{figure}[t]
    \centering \includegraphics[width=0.49\textwidth]{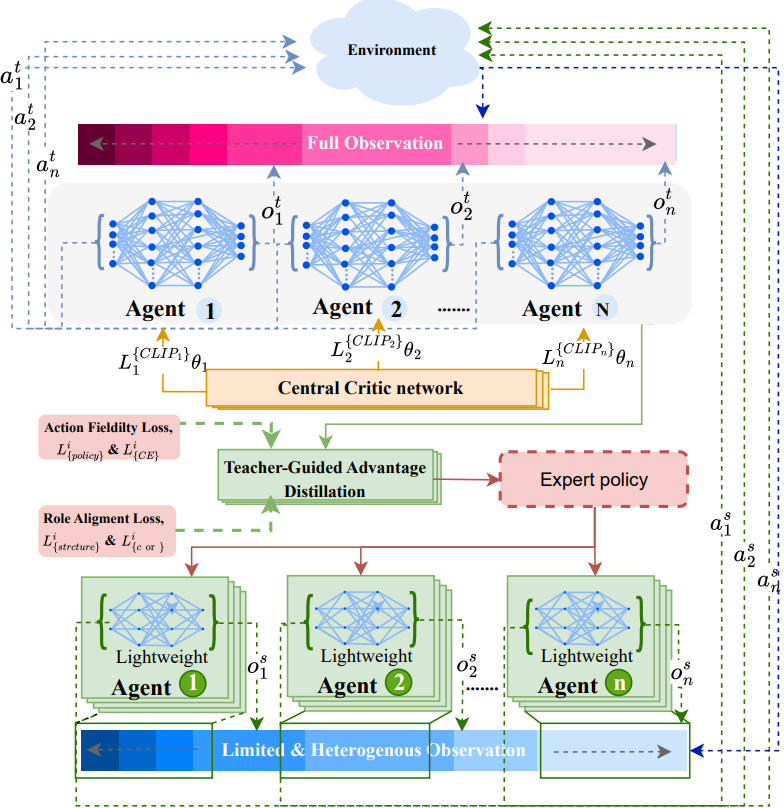}
    \caption{ \small Proposed KD-MARL architecture with two-stage training strategy in limited \& heterogeneous setup.}
    \label{FIG:1}
    \vspace{-4mm}
\end{figure}

In MARL, deploying trained agents in real-world environments presents challenges, particularly in resource-constrained settings due to complex models and large trial-and-error learning cycles, which are computationally expensive. These cycles involve continuous exploration and feedback through interaction with the environment, demanding significant computational resources that make deployment impractical in real-time applications with limited edge resources. To enable deployment in resource-constrained and real-time settings, we propose \textit{KD-MARL} (Fig.~\ref{FIG:1}), a two-stage training framework designed to reduce computational overhead and accelerate decision cycles. In the first stage, a high-capacity teacher policy \(\pi_T\) is trained under centralized training with decentralized execution, using full observations and a centralized critic \(V_T(s)\) to capture joint-state value information. In the second stage, compact student policies \(\pi_S\) are trained without learning any critic. Instead, the critic’s role is replaced through Teacher-Guided Advantage Distillation, where Distilled GAE Advantage Targets computed from the frozen teacher critic provide the policy-gradient signal for student optimization \cite{yu2022surprising}. By eliminating critic inference and reducing network capacity at execution time, the resulting decentralized students achieve faster decision-making and lower computational and memory costs while retaining coordinated near-expert performance.

\subsection{Model Architectures}

Both teacher and student agents utilize recurrent neural network architectures with different hidden dimensions for partially observable multi-agent tasks. Teacher model are large, expressive networks trained offline with state-of-the-art MARL algorithms, capturing long-term dependencies and complex coordination. Student agents adopt structurally similar but smaller architectures with fewer recurrent units, enabling effective knowledge transfer whilst reducing complexity for resource-efficient onboard deployment.

\textbf{Teacher Model.} 
For each agent \(i \in \{1, \dots, N\}\), the teacher employs deep recurrent or feedforward networks with large hidden dimensions (256 or 512 units per layer). Input comprises comprehensive observations \(\mathcal{O}_{T,i} \in \mathbb{R}^{d_{T,i}}\), where \(d_{T,i}\) varies across agents due to heterogeneous roles and features. Output includes action probabilities \(\pi_{T,i}(a|s) \in \Delta^{|\mathcal{A}_i|}\) for policy distillation and value estimates \(V_{T,i}(s) \in \mathbb{R}\), where \(\mathcal{A}_i\) is the agent-specific action space and \(s \in \mathcal{S}\) the joint state.

\textbf{Student Model.} 
Students adopt compact architectures with reduced hidden dimensions (16 or 32) for deployment in resource-limited environments. Each agent \(i\) operates on constrained observations \(\mathcal{O}_{S,i} \in \mathbb{R}^{d_{S,i}}\) where \(d_{S,i} \ll d_{T,i}\). For instance, students access only positional limited features whilst teachers process position, velocity, health, and communication. Observation spaces differ across agents (\(\mathcal{O}_{S,i} \neq \mathcal{O}_{S,j}\)) to recreate heterogeneity tackle uncertainty. Despite reduced and agent-specific inputs, students produce task-consistent action distributions \(\pi_{S,i}(a|s) \in \Delta^{|\mathcal{A}_i|}\) and value estimates \(V_{S,i}(s) \in \mathbb{R}\), ensuring policy fidelity under diverse constraints.

\subsection{Two-stage Training Strategy with Teacher-Guided Advantage Distillation}

The proposed KD-MARL framework adopts a two-stage training strategy (detailed in Algorithm~\ref{alg:student_kd}) to decouple centralized learning from efficient decentralized execution. In the first stage, high-capacity teacher agents are trained using MAPPO under the centralized training and decentralized execution (CTDE) paradigm. Each teacher learns a decentralized policy $\pi_T$ supported by a centralized critic $V_T(s)$, where $s$ denotes the joint environment state. The critic provides variance-reduced value estimates that enable stable learning of coordinated behaviour across agents. After convergence, the teacher policy and critic are frozen and used exclusively to generate supervision signals.

\paragraph{Critic-Free Student Training via Advantage Distillation.}
In the second stage, student agents are trained without learning any critic or value function. Removing the critic directly in actor-critic methods typically leads to unstable policy updates due to high-variance gradients and poor temporal credit assignment. KD-MARL addresses this challenge through \emph{Teacher-Guided Advantage Distillation}, where advantage targets distilled from the frozen teacher critic replace the role of the student critic. This preserves the stabilizing function of the critic while eliminating its computational and memory overhead during student training and execution.

\paragraph{Distilled GAE Advantage Targets.}
When trajectories are sampled from the expert buffer, temporal-difference residuals are computed using the teacher critic $\delta_t^{T}$ through Eq.~\ref{eq_1} where $r_t$ is the shared team reward at time $t$ and $\gamma \in (0,1)$ is the discount factor. These residuals are accumulated using generalized advantage estimation to obtain low-variance, temporally consistent supervision from Eq.~\ref{eq_2}. The resulting distilled advantages $A_T^{\mathrm{GAE}}$ fully replace the student critic and serve as the sole advantage signal during policy optimization.

\begin{equation}
\small
\label{eq_1}
\delta_t^{T} = r_t + \gamma V_T(s_{t+1}) - V_T(s_t),
\end{equation}

\begin{equation}
\small
\label{eq_2}
A_T^{\mathrm{GAE}}(t) = \sum_{l=0}^{L-1} (\gamma \lambda)^l \delta_{t+l}^{T},
\end{equation}


\paragraph{Advantage-Driven Policy optimization.}
Student policies $\pi_S(a_t \mid o_{S,t})$, operating on restricted and heterogeneous local observations $o_{S,t}$, are optimized using a PPO-style objective shown in Eq.(~\ref{PPO_1}) where $r_t(\theta)$ indicates the probability ratio between the updated student policy and the behaviour policy that generated the expert data. Overall, $\epsilon$ constrains policy updates to ensure stable optimization.
\begin{equation}
\label{PPO_1}
r_t(\theta) = \frac{\pi_S(a_t \mid o_{S,t})}{\pi_{\mathrm{beh}}(a_t \mid o_{S,t})}
\end{equation}
\begin{equation}
\label{PPO_2}
\mathcal{L}_{\mathrm{PPO}} =
\mathbb{E}\!\left[
\min\!\left(
\begin{aligned}
& r_t(\theta)\, A_T^{\mathrm{GAE}}(t),\\
& \mathrm{clip}(r_t(\theta), 1-\epsilon, 1+\epsilon)\, A_T^{\mathrm{GAE}}(t)
\end{aligned}
\right)
\right].
\end{equation}

This advantage-guided optimization alone is insufficient to preserve coordinated multi-agent behavior under decentralized execution. To prevent behavioral drift and loss of coordination, policy optimization is regularized using explicit knowledge distillation. The distillation loss $\mathcal{L}_{\mathrm{KD}}$, aligns student and teacher action distributions, preserves inter-agent relational structure, and maintains functional role specialization. The complete objective for Stage~2 student training is shown in Eq.~(\ref{final_one}) where $\mathcal{H}(\pi_S)$ is an entropy regularizer and $\zeta$ controls its contribution. No critic is learned or evaluated during this stage.
\begin{equation}
\label{final_one}
\mathcal{L}_{\mathrm{Stage2}}
=
- \mathcal{L}_{\mathrm{PPO}}
+ \mathcal{L}_{\mathrm{KD}}
- \zeta\, \mathcal{H}(\pi_S)
\end{equation}

In Stage-2, student agents are trained without learning or maintaining any critic or value function. The centralized teacher critic is used only to compute GAE-based advantage targets offline, which replace the critic’s role in policy optimization under a PPO objective with distillation regularization. As a result, the student relies solely on a decentralized actor, producing a lightweight policy suitable for resource-constrained onboard execution.

\subsection{Resource-Aware Deployment.}

 In Stage-2, student agents are trained without learning or maintaining any critic or value function. The centralized teacher critic is used only to compute GAE-based advantage targets offline, which replace the critic’s role in policy optimization under a PPO objective with distillation regularization. In our framework, KD addresses decision-cycle complexity by providing soft targets from teacher distributions, enabling immediate action adjustment without trial-and-error \cite{gou2024reciprocal} and compressing coordination knowledge efficiently \cite{li2024feature,harish2024reinforcement}. As a result, the student relies solely on a decentralized actor, producing a lightweight policy suitable for resource-constrained onboard execution. At deployment, all centralized components, including the teacher critic and expert buffer, are discarded. Execution relies exclusively on ultra-lightweight decentralized student policies operating on local observations. By replacing critic learning with teacher-guided advantage distillation and enforcing coordination preservation through $\mathcal{L}_{\mathrm{KD}}$, KD-MARL achieves stable critic-free training, faster decision cycles, and substantially reduced computational and memory requirements, making it well suited for resource-aware MARL deployment in onboard and edge environments.

\begin{algorithm}[t]
\caption{KD-MARL Student Distillation with Teacher-Guided GAE (Critic-Free)}
\label{alg:student_kd}
\begin{algorithmic}[1]
\REQUIRE Expert buffer $\mathcal{D}_{\text{exp}}$ with $(s,s',r,\{o_T^i,o_S^i,a^i,\text{logits}_T^i,\phi_T^i\}_{i=1}^N)$; frozen teacher $\pi_T$ and critic $V_T$ (full obs, $h_T{=}256$);
students $\{\theta_S^i\}$ (limited hetero obs, $h_S{\in}\{16,32\}$); aligners $\{g_i\}$; role projections $U_T,U_S$; $\gamma,\lambda,\tau,\epsilon$; weights $\alpha,\beta,\lambda_{\text{str}},\lambda_{\text{role}},\zeta$
\STATE \textbf{Stage 1 (Teacher, CTDE):} Train MAPPO teacher $(\pi_T,V_T)$ on full observations; freeze and populate $\mathcal{D}_{\text{exp}}$.
\STATE \textbf{Stage 2 (Student, critic-free):}
\REPEAT
    \STATE Sample minibatch $\mathcal{B}\subset\mathcal{D}_{\text{exp}}$
    \STATE Compute teacher-guided advantages $A_T^{\mathrm{GAE}}$ from $V_T$ \hfill (no student critic)
    \STATE \textbf{for} each agent $i{=}1,\dots,N$ \textbf{do}
        \STATE $\hat{o}^i\!\leftarrow\! g_i(o_S^i)$; $(\ell_S^i,\phi_S^i)\!\leftarrow\!\text{Student}(\hat{o}^i;\theta_S^i)$
        \STATE $\pi_T^i\!\leftarrow\!\mathrm{softmax}(\text{logits}_T^i/\tau)$;\; $\pi_S^i\!\leftarrow\!\mathrm{softmax}(\ell_S^i/\tau)$
        \STATE $\mathcal{L}_{\text{PPO}}^i\!\leftarrow\!\text{PPO-clip}(\pi_S^i,\pi_{\text{beh}}^i,A_T^{\mathrm{GAE}})$ \hfill (critic-free)
        \STATE $\mathcal{L}_{\text{KL}}^i\!\leftarrow\!D_{\text{KL}}(\pi_T^i\parallel \pi_S^i)$;\; $\mathcal{L}_{\text{CE}}^i\!\leftarrow\!\text{CE}(\pi_T^i,\pi_S^i)$
        \STATE $\rho_T^i,\rho_S^i\!\leftarrow\!\mathrm{softmax}(U_T\phi_T^i/\tau),\mathrm{softmax}(U_S\phi_S^i/\tau)$
    \STATE \textbf{end for}
    \STATE $\mathcal{L}_{\text{str}}\!\leftarrow\!\sum_{j<i}\!\big(\cos(\phi_T^i,\phi_T^j)-\cos(\phi_S^i,\phi_S^j)\big)^2$;\;
           $\mathcal{L}_{\text{role}}\!\leftarrow\!\sum_i D_{\text{KL}}(\rho_T^i\parallel\rho_S^i)$
    \STATE $\mathcal{L}\!\leftarrow\!-\sum_i\mathcal{L}_{\text{PPO}}^i+\sum_i(\alpha\mathcal{L}_{\text{KL}}^i+\beta\mathcal{L}_{\text{CE}}^i)+\lambda_{\text{str}}\mathcal{L}_{\text{str}}+\lambda_{\text{role}}\mathcal{L}_{\text{role}}-\zeta\sum_i\mathcal{H}(\pi_S^i)$
    \STATE Update $\{\theta_S^i\}$ and $\{g_i\}$ by gradient descent on $\mathcal{L}$ \hfill (teacher frozen)
\UNTIL{convergence}
\STATE \textbf{return} decentralized students $\{\pi_{\theta_S^i}\}$ for onboard execution (critic discarded)
\end{algorithmic}
\end{algorithm}

\subsection{Distillation Loss}\
\label{policy_dist}
We propose a novel distillation loss function for resource-constrained MARL, enabling ultra-lightweight student agents to inherit both behavioral competence and coordination structure from high-capacity teachers. 
The proposerd distillation loss $\mathcal{L}_{\mathrm{KD}}$ integrates four complementary components, each targeting distinct aspects of expert knowledge transfer with hyperparameters $\alpha$, $\beta$, $\lambda_{\mathrm{structure}}$, and $\lambda_{\mathrm{cor}}$ controlling relative influence.
\vspace{-1mm}
\begin{align}
\small
\mathcal{L}_{\mathrm{KD}}
&= \sum_{i=1}^N \Bigg[
\alpha\, \mathcal{L}_{\mathrm{CE}}^{i} + \mathcal{L}_{\mathrm{CE}}^{i} + \lambda({\mathcal{L}_{\mathrm{strcture}}^{i}+\mathcal{L}_{\mathrm{cor}}^{i})}
\Bigg]
\label{eq:kd_losses}
\end{align}

\subsubsection{Action-Policy Fidelity Based Loss}

To encourage behavioral imitation, we minimize the Kullback–Leibler (KL) divergence between the expert teacher policy and the student policy, using the same policy arguments that appear in the total KD loss. Accordingly, the KL-based action imitation loss for agent $i$ is defined as:
\vspace{-1mm}
\begin{equation}
\small
\mathcal{L}_{\mathrm{policy}}^i
=
\mathbb{E}_{o_t^i \sim d^{\pi^{\mathrm{S}}}}
\left[
D_{\mathrm{KL}}\!\left(
\pi_T^{i}(\cdot \mid o_{T,i})
\;\|\;
\pi_S^{i}(\cdot \mid o_{S,i})
\right)
\right]
\label{eq:policy_distill_KLD}
\end{equation}

This ensures that the student’s action probability distribution remains close to that of the teacher \cite{yang2025multie}.

The second component is the Cross Entropy Loss (Eq. \ref{eq:ce_loss}), which refines the distillation process by
encouraging the student to select the same actions as the
teacher. This loss focuses on aligning the student’s most
probable actions with those of the teacher, ensuring that
the student not only imitates the distribution but also the
teacher’s specific choices \cite{zhang2024entropy}.
\vspace{-1mm}
\begin{equation}
\vspace{-0.5mm}
\small
\begin{aligned}
\mathcal{L}_{\mathrm{CE}}^{i}
&=
\mathrm{CE}\!\big(\pi_T^i(\cdot\mid o_{T,i}),\, \pi_S^i(\cdot\mid o_{S,i})\big)
\\[0.3em]
&=
\mathbb{E}_{o_t^i \sim d^{\pi^{\mathrm{S}}}}
\Bigg[
-
\sum_{a}
\pi_T^i(a \mid o_{T,i})
\log \pi_S^i(a \mid o_{S,i})
\Bigg].
\end{aligned}
\label{eq:ce_loss}
\end{equation}

This loss function $\mathcal{L}_{\mathrm{CE}}^{i}$ ensures that the student agent selects the same action as the teacher for the most probable choices, directly guiding the student to replicate the teacher’s behavior from the expert policy.

\subsubsection{Structure Relation and Coordinated Role based Loss}

While action-level imitation is essential, effective multi-agent distillation additionally requires the preservation of relational geometry and the coordinated role structure encoded by the expert teacher. The expert policy produces latent embeddings $\phi_T^i$ that capture both individual behavioral features and inter-agent dependencies. The student embeddings $\phi_S^i$ must therefore retain these structural properties to support coordinated behavior under restricted observations.

To transfer the teacher's relational geometry, we minimize discrepancies in pairwise cosine similarity between teacher and student latent embeddings. For each agent pair $(i,j)$, the structural relation loss $\mathcal{L}_{\mathrm{structure}}^i$ is defined in Eq. (\ref{eq:structure_loss}) to ensure that student agents maintain consistent relational patterns even with reduced capacity and local observations.

\begin{equation}
\small
\mathcal{L}_{\mathrm{structure}}^i
=
\sum_{j<i}
\left[
\cos\!\big(\phi_T^i,\, \phi_T^j\big)
-
\cos\!\big(\phi_S^i,\, \phi_S^j\big)
\right]^2
\label{eq:structure_loss}
\end{equation}

To further maintain the coordinated behavior learned by the teacher, we include a coordinated role-based loss that aligns the role representations of the teacher and student. Moreover,It ensures that the student's latent role embedding \(\rho_S^i\) remains consistent with the teacher’s role embedding \(\rho_T^i\), preserving role-specific contributions that are essential for cooperative multi-agent decision-making. The coordinated role-based distillation loss is defined in Eq.(~\ref{eq:role_loss}) where the role distributions are computed via $\rho_T^i$ and $\rho_S^i$.

\begin{equation}
\small
\mathcal{L}_{\mathrm{corr}}^i
=
D_{\mathrm{KL}}\!\left(
\rho_T^i \;\|\; \rho_S^i
\right),
\label{eq:role_loss}
\end{equation}

\begin{equation}
\small
\rho_T^i = \mathrm{softmax}\!\left(\frac{U_T \phi_T^i}{\tau}\right),
\qquad
\rho_S^i = \mathrm{softmax}\!\left(\frac{U_S \phi_S^i}{\tau}\right)
\label{eq:rho_teacher_student}
\end{equation}
Here, $U_T$ and $U_S$ denoting the respective role projection matrices, and $\tau$ controlling distribution sharpness. Overall, this coordinated role-based loss $\mathcal{L}_{\mathrm{role}}^i$ prevents the collapse of agent-specific roles during distillation, ensuring that the student agent continues to perform its designated role within the multi-agent system.

\section{Experiments}

\begin{table*}[b]
\centering
\caption{Experimental Results: Algorithm Comparison with KD-MARL on SMAC with Limited \& Heterogeneous observations. All algorithms use three configurations: FO (Full Observation: 109,880 params, hidden dim 256), LH (Limited Heterogeneous: 3,960 params, hidden dim 32), and LH+A (LH with heterogeneous architecture, hidden dim $\in[16,32]$).}
\setlength{\tabcolsep}{1.5pt}
\renewcommand{\arraystretch}{1.15}
\footnotesize
\begin{tabular}{@{}llllccccccccccc@{}}
\toprule
\textbf{Map} & \textbf{Groups} & \textbf{Features} & \textbf{Metric} 
& \multicolumn{3}{c}{\textbf{MAPPO}} 
& \multicolumn{3}{c}{\textbf{QMIX}} 
& \multicolumn{3}{c}{\textbf{VDN}} 
& \multicolumn{2}{c@{}}{\textbf{KD-MARL}} \\
\cmidrule(lr){5-7} \cmidrule(lr){8-10} \cmidrule(lr){11-13} \cmidrule(l){14-15}
 & & & & FO & LH & LH+A & FO & LH & LH+A & FO & LH & LH+A & LH & LH+A \\
\midrule

\multirow{3}{*}{\textbf{3m}} 
& \multirow{3}{*}{\makecell[l]{(0), (1),\\ (2)}} 
& \multirow{3}{*}{\makecell[l]{E, A,\\ A+O}} 
& Return (/20) 
& 19.8$_{\pm0.2}$ & 18.2$_{\pm0.4}$ & 15.0$_{\pm0.7}$ 
& 19.6$_{\pm0.3}$ & 16.0$_{\pm0.6}$ & 12.5$_{\pm0.8}$ 
& 18.0$_{\pm0.5}$ & 13.5$_{\pm0.6}$ & 9.0$_{\pm0.7}$ 
& \textbf{18.6}$_{\pm0.4}$ & \textbf{18.0}$_{\pm0.5}$ \\

& & & Win rate (\%) 
& 98.12 & 92.65 & 80.34 
& \textbf{98.77} & 86.34 & 70.27 
& 85.42 & 68.31 & 52.12 
& 94.78 & 90.39 \\

& & & TPS (ms) 
& 6.5±0.3 & 6.2±0.4 & 6.3±0.4 
& 6.6±0.3 & 5.9±0.4 & 3.8±0.2 
& 6.0±0.3 & 5.4±0.3 & 4.3±0.2 
& 5.5±0.3 & 4.1±0.2 \\

\midrule

\multirow{3}{*}{\textbf{8m}} 
& \multirow{3}{*}{\makecell[l]{(0,1,2), (3,4),\\ (5), (6), (7)}} 
& \multirow{3}{*}{\makecell[l]{E, A, E+O,\\ A+O, A+E}} 
& Return (/20) 
& 17.0$_{\pm1.3}$ & 14.0$_{\pm0.7}$ & 10.0$_{\pm1.2}$ 
& 16.0$_{\pm0.5}$ & 12.5$_{\pm0.9}$ & 8.5$_{\pm1.1}$ 
& 15.0$_{\pm0.8}$ & 10.0$_{\pm0.9}$ & 6.0$_{\pm1.0}$ 
& \textbf{17.8}$_{\pm0.6}$ & \textbf{17.6}$_{\pm0.4}$ \\

& & & Win rate (\%) 
& 89.91 & 77.82 & 60.07 
& \textbf{92.19} & 64.78 & 48.13 
& 75.32 & 52.11 & 33.05 
& 88.97 & 88.23 \\

& & & TPS (ms) 
& 21.5±1.2 & 22.0±1.3 & 21.8±1.2 
& 21.9±1.0 & 18.1±1.1 & 10.8±0.9 
& 19.0±1.0 & 17.2±1.0 & 15.0±0.8 
& 17.3±0.9 & 15.8±0.8 \\

\midrule

\multirow{3}{*}{\textbf{5m\_vs\_6m}} 
& \multirow{3}{*}{\makecell[l]{(0,1), (2),\\ (3), (4)}} 
& \multirow{3}{*}{\makecell[l]{E, E+O+A,\\ A, A+O}} 
& Return (/20) 
& 18.0$_{\pm0.6}$ & 16.5$_{\pm0.5}$ & 13.0$_{\pm0.7}$ 
& \textbf{19.1}$_{\pm0.3}$ & 14.0$_{\pm0.8}$ & 10.0$_{\pm1.0}$ 
& 16.0$_{\pm0.7}$ & 11.0$_{\pm0.8}$ & 7.0$_{\pm0.9}$ 
& 16.8$_{\pm0.5}$ & 16.5$_{\pm0.25}$ \\

& & & Win rate (\%) 
& \textbf{61.85} & 58.09 & 44.78 
& 58.93 & 50.12 & 38.79 
& 50.10 & 38.22 & 25.14 
& 58.66 & 56.15 \\

& & & TPS (ms) 
& 12.0±0.6 & 14.0±1.0 & 12.7±0.7 
& 12.3±0.5 & 10.5±0.6 & 6.2±0.4 
& 11.0±0.5 & 10.2±0.5 & 8.2±0.4 
& 10.0±0.5 & 8.0±0.3 \\

\midrule

\multirow{3}{*}{\textbf{3s5z}} 
& \multirow{3}{*}{\makecell[l]{(0,2), (1),\\ (3,4),(5,6), (7)}} 
& \multirow{3}{*}{\makecell[l]{E, E+A, A,\\ A+O, E+O}} 
& Return (/20) 
& 18.5$_{\pm0.5}$ & 16.8$_{\pm0.5}$ & 13.5$_{\pm0.7}$ 
& \textbf{18.7}$_{\pm0.4}$ & 15.0$_{\pm0.7}$ & 10.5$_{\pm0.9}$ 
& 16.5$_{\pm0.6}$ & 11.0$_{\pm0.7}$ & 7.0$_{\pm0.8}$ 
& 17.2$_{\pm0.5}$ & 16.5$_{\pm0.6}$ \\

& & & Win rate (\%) 
& \textbf{68.31} & 55.66 & 42.54 
& 60.48 & 50.12 & 36.95 
& 53.42 & 40.33 & 24.12 
& 60.28 & 58.17 \\

& & & TPS (ms) 
& 11.5±0.6 & 13.5±0.8 & 12.2±0.7 
& 12.0±0.6 & 10.2±0.6 & 6.0±0.4 
& 10.8±0.6 & 9.8±0.5 & 8.0±0.4 
& 9.7±0.5 & 7.9±0.4 \\

\bottomrule
\end{tabular}
\label{tab:smac_results}
\end{table*}

The experiments were conducted using the EPyMARL and PyMARL2 libraries, which serves as a comprehensive platform for integrating and managing various multi-agent environments. This library enables seamless execution of experiments across the SMAC \& MPE environments, allowing for efficient simulation of agent interactions, reward structures, and learning processes. 

\subsection{Evaluation Environment}
\subsubsection{SMAC}

We first evaluate KD-MARL on the StarCraft Multi-Agent Challenge (SMAC) using the standard maps 3m, 5m\_vs\_6m, 8m, and 3s5z \cite{ellis2023smacv2}. A fully-observant teacher policy is trained to convergence and subsequently used to guide student policies operating under heterogeneous observation constraints. To emulate resource limitations in realistic multi-agent systems, each student agent receives only a subset of feature blocks that includes $O$ (own), $A$ (ally), $E$ (enemy); while the remaining dimensions are masked to zero. For example, in 5m\_vs\_6m, Agents 0-1 observe enemy features only, Agent~2 observes enemy+own+ally, Agent~3 observes ally-only, and Agent~4 observes ally+own. Equivalent masking strategies are consistently applied across other SMAC maps (shown in Table~\ref{tab:smac_results}'s Group and Feature blocks). This setup reflects practical constraints where agents cannot access full situational awareness due to sensing or processing bottlenecks.

\subsubsection{MPE}
We further test KD-MARL in the Multi-Agent Particle Environment (MPE) \cite{lowe2017multi}, where each agent normally observes an 18 dimensional feature vector. To simulate hardware-limited sensing, student agents are restricted to a randomly sampled subset of 8-10 features per episode, with the remaining inputs set to zero. The teacher is trained with full observations, and its policy is distilled to guide constrained students. This design enables the assessment of whether knowledge distillation can effectively transfer expert competence and maintain coordination performance under strict observation budgets.

\subsection{Baselines and Comparisons}

We evaluate KD-MARL against three established multi-agent reinforcement learning baselines: MAPPO \cite{chen2021multiagent}, QMIX \cite{rashid2020monotonic}, and VDN \cite{sunehag2018value}. Experiments are conducted in both the \textit{StarCraft Multi-Agent Challenge (SMAC)} and the \textit{Multi-Agent Particle Environments (MPE)}. Evaluations are carried out under three different settings that vary in the degree of agent heterogeneity and available observations:
\begin{itemize}
    \item \textbf{FO (Full Observation):} All agents have access to the global state or equivalent complete information, representing the ideal coordination scenario.
    \item \textbf{LH (Limited Heterogeneity):} Agents receive only local, role-specific observations, introducing variation in perceptual access and coordination challenges.
    \item \textbf{LH+A (Limited Heterogeneity with Heterogeneous Architectures):} Similar to LH but with agents implemented using different network structures or capacities, simulating deployment on mixed hardware with varying resource constraints.
\end{itemize}

These baselines together cover a spectrum of centralized, partially factorized, and fully decomposed training schemes. The primary goal for comparison is to assess whether lightweight, decentralized agents trained via distillation can retain expert-level behavior with reduced computational and observational resources.

\subsection{Results and Analysis}



\begin{table*}[t]
\centering
\caption{\small Experimental Results: Algorithm Comparison with KD-MARL on MPE with Limited \& Heterogeneous observations. (SL = Speaker-Listerner, SS = Simple Spread, Adv = Adversary, L = Landmark, A = Agent, V = Velocity, P = Position, M = Message, D = Distance).}
\setlength{\tabcolsep}{.7pt}
\renewcommand{\arraystretch}{1}
\footnotesize
\begin{tabular}{@{}llllccccccccccc@{}}
\toprule
\textbf{Map} & \textbf{Groups} & \textbf{Features} & \textbf{Metric} 
& \multicolumn{3}{c}{\textbf{MAPPO}} 
& \multicolumn{3}{c}{\textbf{QMIX}} 
& \multicolumn{3}{c}{\textbf{VDN}} 
& \multicolumn{2}{c@{}}{\textbf{KD-MARL}} \\
\cmidrule(lr){5-7} \cmidrule(lr){8-10} \cmidrule(lr){11-13} \cmidrule(l){14-15}
 & & & 
 & FO & LH & LH+A 
 & FO & LH & LH+A 
 & FO & LH & LH+A 
 & LH & LH+A \\
\midrule

\multirow{2}{*}{\textbf{SL}} 
& \multirow{2}{*}{\makecell[l]{(0),\\ (1)}} 
& \multirow{2}{*}{\makecell[l]{(L, V, P),\\ (A(P), M, V, P)}} 
& Return &
-46.0$\pm$2.0 & -82.0$\pm$3.2 & -118.0$\pm$4.0 
& -55.0$\pm$3.0 & -138.0$\pm$5.5 & -205.0$\pm$7.2 
& -90.0$\pm$4.8 & -170.0$\pm$6.5 & -240.0$\pm$8.0 
& -48.0$\pm$2.2 & -50.0$\pm$2.4 \\

& & & TPS (ms) &
6.0$\pm$0.3 & 5.3$\pm$0.4 & 4.0$\pm$0.3 
& 4.8$\pm$0.3 & 4.6$\pm$0.3 & 4.2$\pm$0.3 
& 4.7$\pm$0.3 & 4.5$\pm$0.3 & 4.3$\pm$0.3 
& 4.0$\pm$0.2 & 3.9$\pm$0.2 \\

\midrule

\multirow{2}{*}{\textbf{SS}} 
& \multirow{2}{*}{\makecell[l]{(0)\\ (1),(2)}} 
& \multirow{2}{*}{\makecell[l]{(L, A(P)),\\ V, P}} 
& Return &
-46.0$\pm$2.1 & -84.0$\pm$3.3 & -120.0$\pm$4.1 
& -55.0$\pm$3.1 & -142.0$\pm$5.8 & -210.0$\pm$7.4 
& -92.0$\pm$4.9 & -185.0$\pm$6.9 & -248.0$\pm$8.6 
& -48.5$\pm$2.1 & -50.5$\pm$2.5 \\

& & & TPS (ms) &
8.1$\pm$0.5 & 7.3$\pm$0.6 & 4.5$\pm$0.4 
& 4.9$\pm$0.3 & 4.6$\pm$0.3 & 4.3$\pm$0.3 
& 5.1$\pm$0.3 & 5.0$\pm$0.3 & 4.7$\pm$0.3 
& 4.2$\pm$0.3 & 4.1$\pm$0.2 \\

\midrule

\multirow{2}{*}{\textbf{Adv}} 
& \multirow{2}{*}{\makecell[l]{(0)\\(1), (2)}} 
& \multirow{2}{*}{\makecell[l]{(L, A(P)),\\(A(P), V), D(A-A)}} 
& Return &
18.0$_{\pm0.6}$ & 16.5$_{\pm0.5}$ & 13.0$_{\pm0.7}$ 
& \textbf{19.1}$_{\pm0.3}$ & 14.0$_{\pm0.8}$ & 10.0$_{\pm1.0}$ 
& 16.0$_{\pm0.7}$ & 11.0$_{\pm0.8}$ & 7.0$_{\pm0.9}$ 
& 16.8$_{\pm0.5}$ & 16.5$_{\pm0.25}$ \\

& & & TPS (ms) &
10.5$\pm$0.7 & 9.4$\pm$0.6 & 6.2$\pm$0.5 
& 7.1$\pm$0.5 & 6.9$\pm$0.5 & 6.2$\pm$0.5 
& 6.2$\pm$0.4 & 6.0$\pm$0.3 & 5.8$\pm$0.4 
& 6.3$\pm$0.4 & 5.9$\pm$0.3 \\

\bottomrule
\end{tabular}
\label{tab:mpe_results}
\vspace{-4mm}
\end{table*}

The experimental results for KD-MARL, compared against MAPPO baseline setup in both SMAC and MPE environments, highlight its ability to preserve expert-level performance while significantly reducing computational overhead.

\paragraph{\textbf{Near-expert accuracy under hetetrogenity.}} KD-MARL performed  close teacher with FO even when policies are compressed and observations are masked. On SMAC \textit{3m}, the LH student attains around $94.0\%$ of MAPPO teacher with FO and maintains a $94.78\%$ win rate with only $3.34\%$ drop. On the harder \textit{8m} task, KD-MARL preserves win rate almost perfectly with $88.97\%$ retention and $0.94\%$ drop, while MAPPO trained directly under LH falls to $77.82\%$ which is $12.09\%$ less from its teacher policy. Similar retention is observed in coordination-heavy settings where on \textit{5m\_vs\_6m}, KD-MARL reache $94.8\%$ win-rate retention, whereas VDN under LH drops to $38.22\%$; on \textit{3s5z}, KD-MARL achieves $60.28\%$, outperforming QMIX-LH and VDN-LH by $+10.16$ and $+19.95$ points, respectively. In MPE, under LH and LH+A, KD-MARL achieves returns within 4-6\% of the FO MAPPO baseline across three maps, while QMIX and VDN suffer substantially larger degradations exceeding 20-40\%. These gaps indicate that, under heterogeneous sensing, distillation primarily mitigates the coordination breakdown that limits non-KD baselines in constraints cases (LH, LH+A).

\begin{figure*}
    \centering
    \begin{overpic}[width=\textwidth]{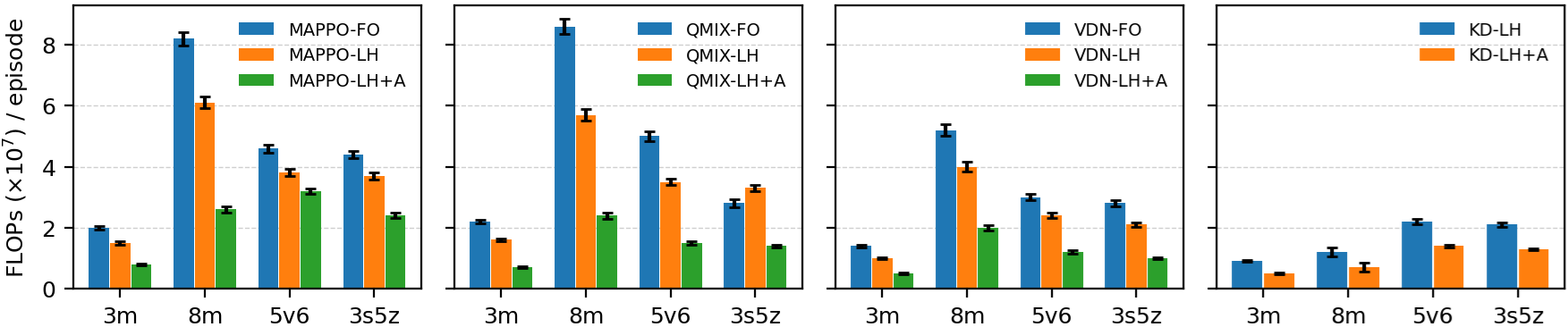}
        \put(2,92){\textbf{(a)}}
        \put(27,92){\textbf{(b)}}
        \put(52,92){\textbf{(c)}}
        \put(77,92){\textbf{(d)}}
    \end{overpic}

    \vspace{2mm}

    \begin{overpic}[width=\textwidth]{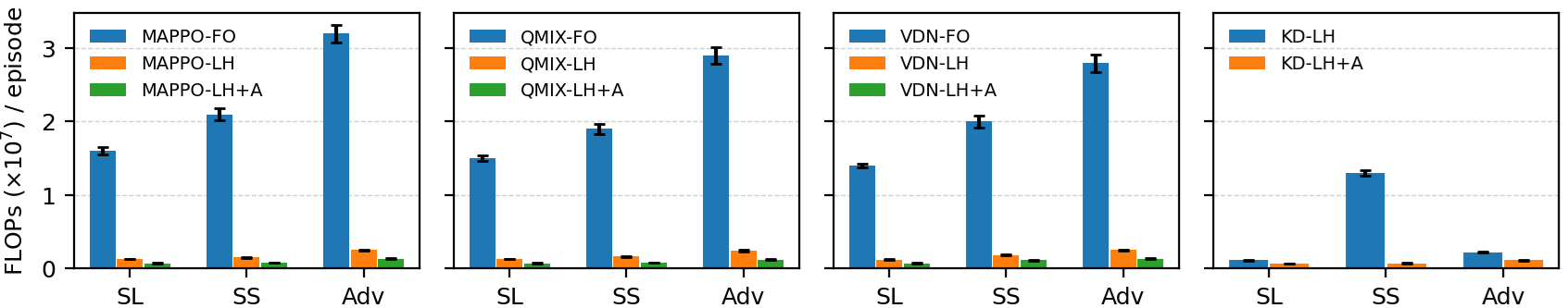}
        \put(2,92){\textbf{(e)}}
        \put(27,92){\textbf{(f)}}
        \put(52,92){\textbf{(g)}}
        \put(77,92){\textbf{(h)}}
    \end{overpic}

    \caption{\small FLOPs per episode comparison across SMAC and MPE maps, demonstrating resource-aware advantages of KD-MARL.
    }
    \label{fig:flops_smac_mpe}
    \vspace{-3mm}
\end{figure*}

\paragraph{\textbf{Resource efficiency.}} The retained performance is achieved with substantially lower compute demonstrating  FLOPs (shown in Fig.~\ref{fig:flops_smac_mpe}) and time per step (TPS) reductions  (shown in Tables~\ref{tab:smac_results} - \ref{tab:mpe_results}). In SMAC, FLOPs reductions range from $3.3\times$ to $11.7\times$ across maps. These savings translate into lower time consumption per step while maintaining near-teacher performance (within $4$--$6\%$ in MPE), whereas QMIX and VDN deteriorate sharply under limited observations, highlighting KD-MARL's suitability for onboard deployment. In MPE, FLOPs are reduced by $26.7\times$ in \textit{Speaker-Listener}, $30.0\times$ in \textit{Simple Spread}, and $29.1\times$ in \textit{Adversary}, yielding an average reduction of approximately $28.6\times$. 
Overall, lightweight computation leads to faster convergence per episode during student model execution with expert policy. In SMAC, runtime falls from $21.5$ to $15.8$ ms per episode on $8m$ with around $26\%$ faster and $33\%$ faster on 5m\_vs\_6m. Nevertheless, KD-MARL achieves this speedup with minimal performance cost, staying within $4$-$6\%$ of the FO teacher on MPE benchmarks. By contrast, QMIX and VDN show steep degradation when observations are restricted, which makes KD-MARL a better fit for deployment scenarios where both speed and coordination quality are essential.

\begin{figure}
    \centering
    \includegraphics[width=0.5\textwidth]{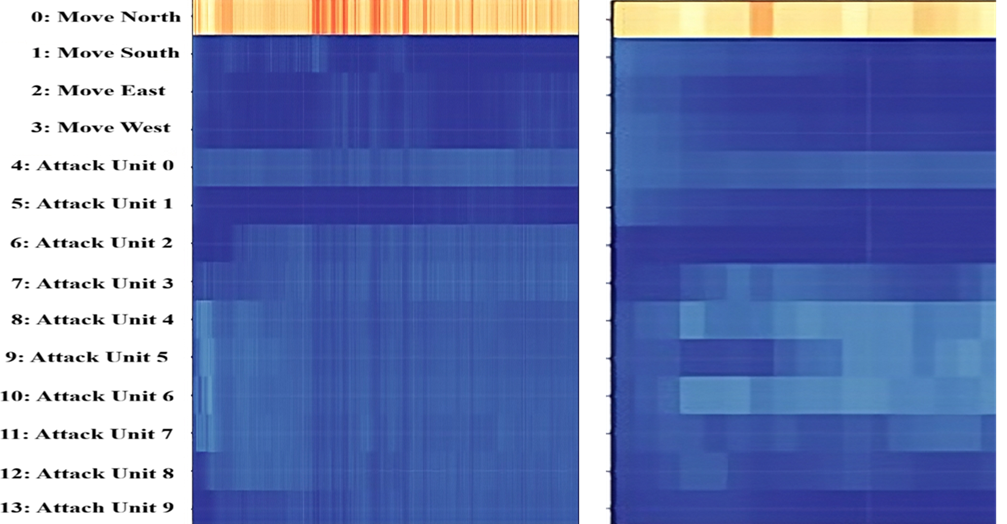}
    \caption{ \small The heatmaps show action-selection frequencies in the 3s5z SMAC scenario under constraints, with non-KD (left) and KD-MARL (right). Warmer colours indicate higher frequency. KD-MARL exhibits more concentrated and stable attack patterns, while the non-KD policy shows more dispersed actions, indicating reduced coordination.}
    \label{fig:heatmaps_3s5z}
\end{figure}

The heatmaps  (in Fig.~\ref{fig:heatmaps_3s5z}) illustrate the action selection frequency over time in the 3s5z StarCraft Multi-Agent Challenge scenario, comparing KD-MARL and non-KD based deployments. The outcomes demonstrate hat the structure relation and role alignment losses preserve inter-agent coordination. In the KD-MARL setup, the action selection, particularly for attack commands (actions 4-13), shows more consistent and coordinated patterns, even under limited and heterogeneous observations. The warmer color intensity in KD-MARL indicates higher frequencies of coordinated actions, especially in attack, compared to the non-KD setup, where coordination is less consistent. This highlights how KD-MARL effectively preserves coordination patterns, ensuring better collective performance despite agent constraints.

\begin{table}[t]
\scriptsize
\centering
\renewcommand{\arraystretch}{1.1}
\caption{\small Comparison of different methods on 3s5z map (SMAC) in heterogeneous setup. (SL = Soft Logit, HL = Hard Logit, IR = Relation Loss, COR = Coordination Loss.)}

\label{tab:comparison}

\begin{tblr}{
  colspec = {Q[160,c] Q[40,c] Q[40,c] Q[40,c] Q[40,c] Q[150,c] Q[100,c] Q[100,c]},
  row{1} = {font=\bfseries, bg=gray!10},
  hline{1,2} = {solid,0.8pt},
  hline{3-Z} = {0.4pt, gray!50},
  vlines = {0.4pt, gray!50},
  rowsep=1pt,
  colsep=2pt,
  stretch=1.2,
}

\textbf{Method} & \textbf{SL} & \textbf{HL} & \textbf{IR} & \textbf{COR} & \textbf{Agent Types} & \textbf{Win Rates (\%)} & \textbf{FLOPs ($\times 10^{7}$)} \\

\cite{zhao2022ctds}     & $\bullet$ & -  &  - &   & RNN             & 46.08 & $1.13$ \\
\cite{tseng2022offline} & $\bullet$ & $\bullet$ & - & -  & Attention+RNN   & 54.92 & $3.05$ \\
\cite{chen2024ptde}     & $\bullet$ & -  & $\bullet$ & -  & Attention+RNN   & 61.56 & $3.17$ \\
\textbf{Ours*}          & $\bullet$ & $\bullet$ & $\bullet$ & $\bullet$ & RNN             & $\textbf{58.17}$ & $1.30 \pm 0.03$ \\

\end{tblr}
\vspace{-6mm}
\end{table}

Table~\ref{tab:comparison} presents broad comparison of performance retention across different approaches, showing win rates and FLOPs for various methods. Here, the performance retention is evaluated based on win percentages, where higher values indicate better retention of expert performance. Although other methods such as \cite{zhao2022ctds} and \cite{tseng2022offline} show higher win rates, our approach demonstrates a better balance between performance retention and computational efficiency. Specifically, while methods like \cite{chen2024ptde} achieve higher win rates, they also incur significantly higher computational costs in terms of FLOPs. In contrast, our method achieves a competitive win rate of around 58.17\%, with 1.3 ± 0.03($\times10^{7}$) FLOPs, demonstrating superior performance retention with lower computational overhead. This confirms that our approach effectively balances both performance and resource efficiency, making it more suitable for real-world applications where computational resources are limited.





\section{Conclusion}
This work presented Resource-Aware Knowledge Distillation based Multi-Agent Reinforcement Learning (KD-MARL), a comprehensive framework for achieving efficient and coordinated decision-making under strict computational and observational constraints. The study contributes (i) a two-stage distillation paradigm that transfers coordination knowledge from a high-capacity expert to ultra-lightweight, heterogeneous student agents,(ii) critic-free student optimization strategy based on distilled advantage signals and structured policy distillation, and (iii) extensive empirical validation across SMAC and MPE benchmarks. KD-MARL achieves near-expert performance retention of \(92\%\), while reducing FLOPs by up to \(96.5\%\) and inference time by approximately \(40\%\), confirming its capacity for real-time, decentralized execution in resource-limited systems. most of the teacher's performance, staying within $4$--$6\%$ in MPE and retaining over $90\%$ effectiveness across SMAC scenarios, while delivering major efficiency improvements. Specifically, we observe FLOPs reductions of up to $28.6\times$ in MPE and between $3.3\times$ and $11.7\times$ in SMAC, with throughput gains reaching $48\%$ and $33\%$ respectively. These findings indicate that KD-MARL successfully balances coordination quality with computational efficiency, making it practical for real-time deployment on resource-limited platforms. Future extensions will focus on online adaptive distillation, multi-teacher transfer, and communication-efficient coordination, enabling broader applicability to autonomous and edge-level multi-agent systems.


\section*{Acknowledgments}

\noindent This work has been supported by the SmartSat CRC, whose activities are funded by the Australian Government’s CRC Program. This work use an open-source realistic satellite simulator (Basilisk and BSK-RL) that is actively developed by Dr. Hanspeter Schaub and team at AVS Laboratory, University of Colorado Boulder. Also, the authors would like to express their sincere gratitude to BAE Systems for their invaluable support and collaboration throughout this research.



{\small\setlength{\itemsep}{0pt}
\bibliographystyle{cas-model2-names}
{\footnotesize
\setlength{\itemsep}{0pt}
\setlength{\parskip}{0pt}
\setlength{\bibsep}{0pt}
\bibliography{cas-refs}

\begin{thebibliography}{37}
\expandafter\ifx\csname natexlab\endcsname\relax\def\natexlab#1{#1}\fi
\providecommand{\url}[1]{\texttt{#1}}
\providecommand{\href}[2]{#2}
\providecommand{\path}[1]{#1}
\providecommand{\DOIprefix}{doi:}
\providecommand{\ArXivprefix}{arXiv:}
\providecommand{\URLprefix}{URL: }
\providecommand{\Pubmedprefix}{pmid:}
\providecommand{\doi}[1]{\href{http://dx.doi.org/#1}{\path{#1}}}
\providecommand{\Pubmed}[1]{\href{pmid:#1}{\path{#1}}}
\providecommand{\bibinfo}[2]{#2}
\ifx\xfnm\relax \def\xfnm[#1]{\unskip,\space#1}\fi
\bibitem[{Bao et~al.(2022)Bao, Ma and Yi}]{bao2022recent}
\bibinfo{author}{Bao, G.}, \bibinfo{author}{Ma, L.}, \bibinfo{author}{Yi, X.}, \bibinfo{year}{2022}.
\newblock \bibinfo{title}{Recent advances on cooperative control of heterogeneous multi-agent systems subject to constraints: A survey}.
\newblock \bibinfo{journal}{Systems Science \& Control Engineering} \bibinfo{volume}{10}, \bibinfo{pages}{539--551}.
\bibitem[{Chen et~al.(2021)Chen, Hu, Guan, Zhao and Shen}]{chen2021multiagent}
\bibinfo{author}{Chen, L.}, \bibinfo{author}{Hu, B.}, \bibinfo{author}{Guan, Z.H.}, \bibinfo{author}{Zhao, L.}, \bibinfo{author}{Shen, X.}, \bibinfo{year}{2021}.
\newblock \bibinfo{title}{Multiagent meta-reinforcement learning for adaptive multipath routing optimization}.
\newblock \bibinfo{journal}{IEEE Transactions on Neural Networks and Learning Systems} \bibinfo{volume}{33}, \bibinfo{pages}{5374--5386}.
\bibitem[{Chen et~al.(2023)Chen, Lin and Zhou}]{chen2023portfolio}
\bibinfo{author}{Chen, R.}, \bibinfo{author}{Lin, J.}, \bibinfo{author}{Zhou, Y.}, \bibinfo{year}{2023}.
\newblock \bibinfo{title}{Portfolio management with multi-agent reinforcement learning: A role-aware distillation approach}.
\newblock \bibinfo{journal}{Journal of Financial Data Science} .
\bibitem[{Chen et~al.(2024)Chen, Mao, Mao, Wu, Zhang, Zhang, Yang and Chang}]{chen2024ptde}
\bibinfo{author}{Chen, Y.}, \bibinfo{author}{Mao, H.}, \bibinfo{author}{Mao, J.}, \bibinfo{author}{Wu, S.}, \bibinfo{author}{Zhang, T.}, \bibinfo{author}{Zhang, B.}, \bibinfo{author}{Yang, W.}, \bibinfo{author}{Chang, H.}, \bibinfo{year}{2024}.
\newblock \bibinfo{title}{Ptde: personalized training with distilled execution for multi-agent reinforcement learning}, in: \bibinfo{booktitle}{Proceedings of the Thirty-Third International Joint Conference on Artificial Intelligence}, pp. \bibinfo{pages}{31--39}.
\bibitem[{Czarnecki et~al.(2019)Czarnecki, Pascanu, Osindero, Jayakumar, Swirszcz and Jaderberg}]{czarnecki2019distilling}
\bibinfo{author}{Czarnecki, W.M.}, \bibinfo{author}{Pascanu, R.}, \bibinfo{author}{Osindero, S.}, \bibinfo{author}{Jayakumar, S.}, \bibinfo{author}{Swirszcz, G.}, \bibinfo{author}{Jaderberg, M.}, \bibinfo{year}{2019}.
\newblock \bibinfo{title}{Distilling policy distillation}, in: \bibinfo{booktitle}{The 22nd international conference on artificial intelligence and statistics}, \bibinfo{organization}{PMLR}. pp. \bibinfo{pages}{1331--1340}.
\bibitem[{Dan et~al.(2024)Dan, Wang and He}]{dan2024pdd}
\bibinfo{author}{Dan, X.}, \bibinfo{author}{Wang, L.}, \bibinfo{author}{He, Z.}, \bibinfo{year}{2024}.
\newblock \bibinfo{title}{Pdd: Pruning during distillation for efficient multi-agent reinforcement learning}, in: \bibinfo{booktitle}{Proceedings of the AAAI Conference on Artificial Intelligence}.
\bibitem[{De~Nijs et~al.(2021)De~Nijs, Walraven, De~Weerdt and Spaan}]{de2021constrained}
\bibinfo{author}{De~Nijs, F.}, \bibinfo{author}{Walraven, E.}, \bibinfo{author}{De~Weerdt, M.}, \bibinfo{author}{Spaan, M.}, \bibinfo{year}{2021}.
\newblock \bibinfo{title}{Constrained multiagent markov decision processes: A taxonomy of problems and algorithms}.
\newblock \bibinfo{journal}{Journal of Artificial Intelligence Research} \bibinfo{volume}{70}, \bibinfo{pages}{955--1001}.
\bibitem[{Ellis et~al.(2023)Ellis, Cook, Moalla, Samvelyan, Sun, Mahajan, Foerster and Whiteson}]{ellis2023smacv2}
\bibinfo{author}{Ellis, B.}, \bibinfo{author}{Cook, J.}, \bibinfo{author}{Moalla, S.}, \bibinfo{author}{Samvelyan, M.}, \bibinfo{author}{Sun, M.}, \bibinfo{author}{Mahajan, A.}, \bibinfo{author}{Foerster, J.}, \bibinfo{author}{Whiteson, S.}, \bibinfo{year}{2023}.
\newblock \bibinfo{title}{Smacv2: An improved benchmark for cooperative multi-agent reinforcement learning}.
\newblock \bibinfo{journal}{Advances in Neural Information Processing Systems} \bibinfo{volume}{36}, \bibinfo{pages}{37567--37593}.
\bibitem[{Foerster et~al.(2016)Foerster, Assael, De~Freitas and Whiteson}]{foerster2016learning}
\bibinfo{author}{Foerster, J.}, \bibinfo{author}{Assael, I.A.}, \bibinfo{author}{De~Freitas, N.}, \bibinfo{author}{Whiteson, S.}, \bibinfo{year}{2016}.
\newblock \bibinfo{title}{Learning to communicate with deep multi-agent reinforcement learning}.
\newblock \bibinfo{journal}{Advances in neural information processing systems} \bibinfo{volume}{29}.
\bibitem[{Gao et~al.(2021)Gao, Zhang, Yang, Li, Li and Hu}]{gao2021knowru}
\bibinfo{author}{Gao, Y.}, \bibinfo{author}{Zhang, K.}, \bibinfo{author}{Yang, Y.}, \bibinfo{author}{Li, Y.}, \bibinfo{author}{Li, Z.}, \bibinfo{author}{Hu, H.}, \bibinfo{year}{2021}.
\newblock \bibinfo{title}{Knowru: Knowledge reuse in multi-agent reinforcement learning}.
\newblock \bibinfo{journal}{Neurocomputing} \bibinfo{volume}{453}, \bibinfo{pages}{464--475}.
\bibitem[{Gou et~al.(2024)Gou, Chen, Yu, Liu, Du, Wan and Yi}]{gou2024reciprocal}
\bibinfo{author}{Gou, J.}, \bibinfo{author}{Chen, Y.}, \bibinfo{author}{Yu, B.}, \bibinfo{author}{Liu, J.}, \bibinfo{author}{Du, L.}, \bibinfo{author}{Wan, S.}, \bibinfo{author}{Yi, Z.}, \bibinfo{year}{2024}.
\newblock \bibinfo{title}{Reciprocal teacher-student learning via forward and feedback knowledge distillation}.
\newblock \bibinfo{journal}{IEEE transactions on multimedia} \bibinfo{volume}{26}, \bibinfo{pages}{7901--7916}.
\bibitem[{Gronauer and Diepold(2022)}]{gronauer2022multi}
\bibinfo{author}{Gronauer, S.}, \bibinfo{author}{Diepold, K.}, \bibinfo{year}{2022}.
\newblock \bibinfo{title}{Multi-agent deep reinforcement learning: a survey}.
\newblock \bibinfo{journal}{Artificial Intelligence Review} \bibinfo{volume}{55}, \bibinfo{pages}{895--943}.
\bibitem[{Harish et~al.(2024)Harish, Heck, Hanna, Kira and Szot}]{harish2024reinforcement}
\bibinfo{author}{Harish, A.N.}, \bibinfo{author}{Heck, L.}, \bibinfo{author}{Hanna, J.P.}, \bibinfo{author}{Kira, Z.}, \bibinfo{author}{Szot, A.}, \bibinfo{year}{2024}.
\newblock \bibinfo{title}{Reinforcement learning via auxiliary task distillation}, in: \bibinfo{booktitle}{European Conference on Computer Vision}, \bibinfo{organization}{Springer}. pp. \bibinfo{pages}{214--230}.
\bibitem[{Hu et~al.(2023)Hu, Li, Liu, Wu, Chen, Wang and Liu}]{hu2023teacher}
\bibinfo{author}{Hu, C.}, \bibinfo{author}{Li, X.}, \bibinfo{author}{Liu, D.}, \bibinfo{author}{Wu, H.}, \bibinfo{author}{Chen, X.}, \bibinfo{author}{Wang, J.}, \bibinfo{author}{Liu, X.}, \bibinfo{year}{2023}.
\newblock \bibinfo{title}{Teacher-student architecture for knowledge distillation: A survey}.
\newblock \bibinfo{journal}{arXiv preprint arXiv:2308.04268} .
\bibitem[{Hu et~al.(2024)Hu, Li, Li, Pan and Huang}]{hu2024value}
\bibinfo{author}{Hu, P.}, \bibinfo{author}{Li, S.}, \bibinfo{author}{Li, Z.}, \bibinfo{author}{Pan, L.}, \bibinfo{author}{Huang, L.}, \bibinfo{year}{2024}.
\newblock \bibinfo{title}{Value-based deep multi-agent reinforcement learning with dynamic sparse training}.
\newblock \bibinfo{journal}{arXiv preprint arXiv:2409.19391} .
\bibitem[{Jiang et~al.(2019)Jiang, Feng, Qin and Liu}]{jiang2019multi}
\bibinfo{author}{Jiang, W.}, \bibinfo{author}{Feng, G.}, \bibinfo{author}{Qin, S.}, \bibinfo{author}{Liu, Y.}, \bibinfo{year}{2019}.
\newblock \bibinfo{title}{Multi-agent reinforcement learning based cooperative content caching for mobile edge networks}.
\newblock \bibinfo{journal}{IEEE Access} \bibinfo{volume}{7}, \bibinfo{pages}{61856--61867}.
\bibitem[{Li et~al.(2025)Li, Hu and Tang}]{ddn2025}
\bibinfo{author}{Li, Z.}, \bibinfo{author}{Hu, X.}, \bibinfo{author}{Tang, J.}, \bibinfo{year}{2025}.
\newblock \bibinfo{title}{Double distillation network for robust multi-agent coordination}.
\newblock \bibinfo{journal}{IEEE Transactions on Pattern Analysis and Machine Intelligence} \bibinfo{note}{In press}.
\bibitem[{Li et~al.(2024)Li, Xu, Dong, Zhang and Deng}]{li2024feature}
\bibinfo{author}{Li, Z.}, \bibinfo{author}{Xu, P.}, \bibinfo{author}{Dong, Z.}, \bibinfo{author}{Zhang, R.}, \bibinfo{author}{Deng, Z.}, \bibinfo{year}{2024}.
\newblock \bibinfo{title}{Feature-level knowledge distillation for place recognition based on soft-hard labels teaching paradigm}.
\newblock \bibinfo{journal}{IEEE Transactions on Intelligent Transportation Systems} .
\bibitem[{Liu et~al.(2025)Liu, Zhu, Liu, Liu, Han, Tian, Li and Yi}]{liu2025survey}
\bibinfo{author}{Liu, D.}, \bibinfo{author}{Zhu, Y.}, \bibinfo{author}{Liu, Z.}, \bibinfo{author}{Liu, Y.}, \bibinfo{author}{Han, C.}, \bibinfo{author}{Tian, J.}, \bibinfo{author}{Li, R.}, \bibinfo{author}{Yi, W.}, \bibinfo{year}{2025}.
\newblock \bibinfo{title}{A survey of model compression techniques: Past, present, and future}.
\newblock \bibinfo{journal}{Frontiers in Robotics and AI} \bibinfo{volume}{12}, \bibinfo{pages}{1518965}.
\bibitem[{Liu et~al.(2024)Liu, Huang, Wang, Gao and Chen}]{liu2024fine}
\bibinfo{author}{Liu, K.}, \bibinfo{author}{Huang, Z.}, \bibinfo{author}{Wang, C.D.}, \bibinfo{author}{Gao, B.}, \bibinfo{author}{Chen, Y.}, \bibinfo{year}{2024}.
\newblock \bibinfo{title}{Fine-grained learning behavior-oriented knowledge distillation for graph neural networks}.
\newblock \bibinfo{journal}{IEEE Transactions on Neural Networks and Learning Systems} .
\bibitem[{Liu et~al.(2023)Liu, Chen and Zhang}]{liu2023modelcompression}
\bibinfo{author}{Liu, W.}, \bibinfo{author}{Chen, J.}, \bibinfo{author}{Zhang, M.}, \bibinfo{year}{2023}.
\newblock \bibinfo{title}{Model compression in multi-agent reinforcement learning via reinforcement learning-guided pruning}.
\newblock \bibinfo{journal}{IEEE Transactions on Neural Networks and Learning Systems} \bibinfo{note}{To appear}.
\bibitem[{Lowe et~al.(2017)Lowe, Wu, Tamar, Harb, Pieter~Abbeel and Mordatch}]{lowe2017multi}
\bibinfo{author}{Lowe, R.}, \bibinfo{author}{Wu, Y.I.}, \bibinfo{author}{Tamar, A.}, \bibinfo{author}{Harb, J.}, \bibinfo{author}{Pieter~Abbeel, O.}, \bibinfo{author}{Mordatch, I.}, \bibinfo{year}{2017}.
\newblock \bibinfo{title}{Multi-agent actor-critic for mixed cooperative-competitive environments}.
\newblock \bibinfo{journal}{Advances in neural information processing systems} \bibinfo{volume}{30}.
\bibitem[{Nekoei et~al.(2023)Nekoei, Badrinaaraayanan, Sinha, Amini, Rajendran, Mahajan and Chandar}]{nekoei2023dealing}
\bibinfo{author}{Nekoei, H.}, \bibinfo{author}{Badrinaaraayanan, A.}, \bibinfo{author}{Sinha, A.}, \bibinfo{author}{Amini, M.}, \bibinfo{author}{Rajendran, J.}, \bibinfo{author}{Mahajan, A.}, \bibinfo{author}{Chandar, S.}, \bibinfo{year}{2023}.
\newblock \bibinfo{title}{Dealing with non-stationarity in decentralized cooperative multi-agent deep reinforcement learning via multi-timescale learning}, in: \bibinfo{booktitle}{Conference on Lifelong Learning Agents}, \bibinfo{organization}{PMLR}. pp. \bibinfo{pages}{376--398}.
\bibitem[{Park et~al.(2019)Park, Kim, Lu and Cho}]{park2019relational}
\bibinfo{author}{Park, W.}, \bibinfo{author}{Kim, D.}, \bibinfo{author}{Lu, Y.}, \bibinfo{author}{Cho, M.}, \bibinfo{year}{2019}.
\newblock \bibinfo{title}{Relational knowledge distillation}, in: \bibinfo{booktitle}{Proceedings of the IEEE/CVF conference on computer vision and pattern recognition}, pp. \bibinfo{pages}{3967--3976}.
\bibitem[{Pei et~al.(2025)Pei, Ren, Zhang, Sun and Champeyrol}]{pei2025policy}
\bibinfo{author}{Pei, Y.}, \bibinfo{author}{Ren, T.}, \bibinfo{author}{Zhang, Y.}, \bibinfo{author}{Sun, Z.}, \bibinfo{author}{Champeyrol, M.}, \bibinfo{year}{2025}.
\newblock \bibinfo{title}{Policy distillation for efficient decentralized execution in multi-agent reinforcement learning}.
\newblock \bibinfo{journal}{Neurocomputing} , \bibinfo{pages}{129617}.
\bibitem[{Rashid et~al.(2020)Rashid, Samvelyan, De~Witt, Farquhar, Foerster and Whiteson}]{rashid2020monotonic}
\bibinfo{author}{Rashid, T.}, \bibinfo{author}{Samvelyan, M.}, \bibinfo{author}{De~Witt, C.S.}, \bibinfo{author}{Farquhar, G.}, \bibinfo{author}{Foerster, J.}, \bibinfo{author}{Whiteson, S.}, \bibinfo{year}{2020}.
\newblock \bibinfo{title}{Monotonic value function factorisation for deep multi-agent reinforcement learning}.
\newblock \bibinfo{journal}{Journal of Machine Learning Research} \bibinfo{volume}{21}, \bibinfo{pages}{1--51}.
\bibitem[{Sunehag et~al.(2018)Sunehag, Lever, Gruslys, Czarnecki, Zambaldi, Jaderberg, Lanctot, Sonnerat, Leibo, Tuyls et~al.}]{sunehag2018value}
\bibinfo{author}{Sunehag, P.}, \bibinfo{author}{Lever, G.}, \bibinfo{author}{Gruslys, A.}, \bibinfo{author}{Czarnecki, W.M.}, \bibinfo{author}{Zambaldi, V.}, \bibinfo{author}{Jaderberg, M.}, \bibinfo{author}{Lanctot, M.}, \bibinfo{author}{Sonnerat, N.}, \bibinfo{author}{Leibo, J.Z.}, \bibinfo{author}{Tuyls, K.}, et~al., \bibinfo{year}{2018}.
\newblock \bibinfo{title}{Value-decomposition networks for cooperative multi-agent learning based on team reward}, in: \bibinfo{booktitle}{Proceedings of the 17th International Conference on Autonomous Agents and MultiAgent Systems}, pp. \bibinfo{pages}{2085--2087}.
\bibitem[{Tseng et~al.(2022)Tseng, Wang, Lin and Isola}]{tseng2022offline}
\bibinfo{author}{Tseng, W.C.}, \bibinfo{author}{Wang, T.H.J.}, \bibinfo{author}{Lin, Y.C.}, \bibinfo{author}{Isola, P.}, \bibinfo{year}{2022}.
\newblock \bibinfo{title}{Offline multi-agent reinforcement learning with knowledge distillation}.
\newblock \bibinfo{journal}{Advances in Neural Information Processing Systems} \bibinfo{volume}{35}, \bibinfo{pages}{226--237}.
\bibitem[{Wang et~al.(2022)Wang, Zhao and Liu}]{offline2022}
\bibinfo{author}{Wang, X.}, \bibinfo{author}{Zhao, Y.}, \bibinfo{author}{Liu, Q.}, \bibinfo{year}{2022}.
\newblock \bibinfo{title}{Offline multi-agent reinforcement learning via knowledge distillation}, in: \bibinfo{booktitle}{Advances in Neural Information Processing Systems (NeurIPS)}.
\bibitem[{Wong et~al.(2023)Wong, B{\"a}ck, Kononova and Plaat}]{wong2023deep}
\bibinfo{author}{Wong, A.}, \bibinfo{author}{B{\"a}ck, T.}, \bibinfo{author}{Kononova, A.V.}, \bibinfo{author}{Plaat, A.}, \bibinfo{year}{2023}.
\newblock \bibinfo{title}{Deep multiagent reinforcement learning: Challenges and directions}.
\newblock \bibinfo{journal}{Artificial Intelligence Review} \bibinfo{volume}{56}, \bibinfo{pages}{5023--5056}.
\bibitem[{Xu et~al.(2025)Xu, Wang, Xu, Yu, Huang and Yi}]{xu2025survey}
\bibinfo{author}{Xu, Z.}, \bibinfo{author}{Wang, J.}, \bibinfo{author}{Xu, X.}, \bibinfo{author}{Yu, P.}, \bibinfo{author}{Huang, T.}, \bibinfo{author}{Yi, J.}, \bibinfo{year}{2025}.
\newblock \bibinfo{title}{A survey of reinforcement learning-driven knowledge distillation: Techniques, challenges, and applications} .
\bibitem[{Yang et~al.(2025)Yang, Yu, Yang, An, Yu, Huang and Xu}]{yang2025multi}
\bibinfo{author}{Yang, C.}, \bibinfo{author}{Yu, X.}, \bibinfo{author}{Yang, H.}, \bibinfo{author}{An, Z.}, \bibinfo{author}{Yu, C.}, \bibinfo{author}{Huang, L.}, \bibinfo{author}{Xu, Y.}, \bibinfo{year}{2025}.
\newblock \bibinfo{title}{Multi-teacher knowledge distillation with reinforcement learning for visual recognition}, in: \bibinfo{booktitle}{Proceedings of the AAAI Conference on Artificial Intelligence}, pp. \bibinfo{pages}{9148--9156}.
\bibitem[{Yang et~al.(2024)Yang, Chen, Zhang and Berry}]{yang2024beyond}
\bibinfo{author}{Yang, N.}, \bibinfo{author}{Chen, S.}, \bibinfo{author}{Zhang, H.}, \bibinfo{author}{Berry, R.}, \bibinfo{year}{2024}.
\newblock \bibinfo{title}{Beyond the edge: An advanced exploration of reinforcement learning for mobile edge computing, its applications, and future research trajectories}.
\newblock \bibinfo{journal}{IEEE Communications Surveys \& Tutorials} \bibinfo{volume}{27}, \bibinfo{pages}{546--594}.
\bibitem[{Yu et~al.(2022)Yu, Velu, Vinitsky, Gao, Wang, Bayen and Wu}]{yu2022surprising}
\bibinfo{author}{Yu, C.}, \bibinfo{author}{Velu, A.}, \bibinfo{author}{Vinitsky, E.}, \bibinfo{author}{Gao, J.}, \bibinfo{author}{Wang, Y.}, \bibinfo{author}{Bayen, A.}, \bibinfo{author}{Wu, Y.}, \bibinfo{year}{2022}.
\newblock \bibinfo{title}{The surprising effectiveness of ppo in cooperative multi-agent games}.
\newblock \bibinfo{journal}{Advances in neural information processing systems} \bibinfo{volume}{35}, \bibinfo{pages}{24611--24624}.
\bibitem[{Zhang et~al.(2024)Zhang, Luo, Sj{\"o}lund, Sch{\"o}n and Mattsson}]{zhang2024entropy}
\bibinfo{author}{Zhang, R.}, \bibinfo{author}{Luo, Z.}, \bibinfo{author}{Sj{\"o}lund, J.}, \bibinfo{author}{Sch{\"o}n, T.}, \bibinfo{author}{Mattsson, P.}, \bibinfo{year}{2024}.
\newblock \bibinfo{title}{Entropy-regularized diffusion policy with q-ensembles for offline reinforcement learning}.
\newblock \bibinfo{journal}{Advances in Neural Information Processing Systems} \bibinfo{volume}{37}, \bibinfo{pages}{98871--98897}.
\bibitem[{Zhao et~al.(2022)Zhao, Hu, Yang, Zhou, Zhu and Li}]{zhao2022ctds}
\bibinfo{author}{Zhao, J.}, \bibinfo{author}{Hu, X.}, \bibinfo{author}{Yang, M.}, \bibinfo{author}{Zhou, W.}, \bibinfo{author}{Zhu, J.}, \bibinfo{author}{Li, H.}, \bibinfo{year}{2022}.
\newblock \bibinfo{title}{Ctds: Centralized teacher with decentralized student for multiagent reinforcement learning}.
\newblock \bibinfo{journal}{IEEE Transactions on Games} \bibinfo{volume}{16}, \bibinfo{pages}{140--150}.
\bibitem[{Zhong et~al.(2024)Zhong, Kuba, Feng, Hu, Ji and Yang}]{zhong2024heterogeneous}
\bibinfo{author}{Zhong, Y.}, \bibinfo{author}{Kuba, J.G.}, \bibinfo{author}{Feng, X.}, \bibinfo{author}{Hu, S.}, \bibinfo{author}{Ji, J.}, \bibinfo{author}{Yang, Y.}, \bibinfo{year}{2024}.
\newblock \bibinfo{title}{Heterogeneous-agent reinforcement learning}.
\newblock \bibinfo{journal}{Journal of Machine Learning Research} \bibinfo{volume}{25}, \bibinfo{pages}{1--67}.

\end{thebibliography}
}
}

\end{document}